\RequirePackage{fix-cm}
\documentclass[twocolumn] {svjour3}          
\smartqed  
\usepackage{graphicx}
\usepackage{breakcites}
\usepackage{flushend}
\usepackage{amsmath,amssymb} 
\usepackage[colorlinks,citecolor=blue,urlcolor=red,bookmarks=false,hypertexnames=true]{hyperref} 
\usepackage{dblfloatfix}
\usepackage{booktabs}

\DeclareMathDelimiter{\lVert}
  {\mathopen}{symbols}{"6B}{largesymbols}{"0D}
\DeclareMathDelimiter{\rVert}
  {\mathclose}{symbols}{"6B}{largesymbols}{"0D}
\begin{document}
\sloppy
\title{All About Knowledge Graphs for Actions
}


\author{Pallabi Ghosh$^1$        \and
        Nirat Saini$^1$\and
        Larry S. Davis$^1$ \and
        Abhinav Shrivastava$^1$
}
\institute{$^1$Department of Computer Science, University of Maryland, College Park, MD, USA\\
\email{\{pallabig, nirat, lsd, abhinav\}@cs.umd.edu}}

\authorrunning{Ghosh et al.} 



\maketitle

\begin{abstract}
Current action recognition systems require large amounts of training data for recognizing an action. Recent works have explored the paradigm of zero-shot and few-shot learning to learn classifiers for unseen categories or categories with few labels. Following similar paradigms in object recognition, these approaches utilize external sources of knowledge (eg. knowledge graphs from language domains). However, unlike objects, it is unclear what is the best knowledge representation for actions. In this paper, we intend to gain a better understanding of knowledge graphs (KGs) that can be utilized for zero-shot and few-shot action recognition. In particular, we study three different construction mechanisms for KGs: action embeddings, action-object embeddings, visual embeddings. We present extensive analysis of the impact of different KGs in different experimental setups. Finally, to enable a systematic study of zero-shot and few-shot approaches, we propose an improved evaluation paradigm based on UCF101, HMDB51, and Charades datasets for knowledge transfer from models trained on Kinetics.
\end{abstract}







\section{Introduction}
\label{intro}
\begin{figure*}
    \centering
\includegraphics[width=0.9\linewidth]{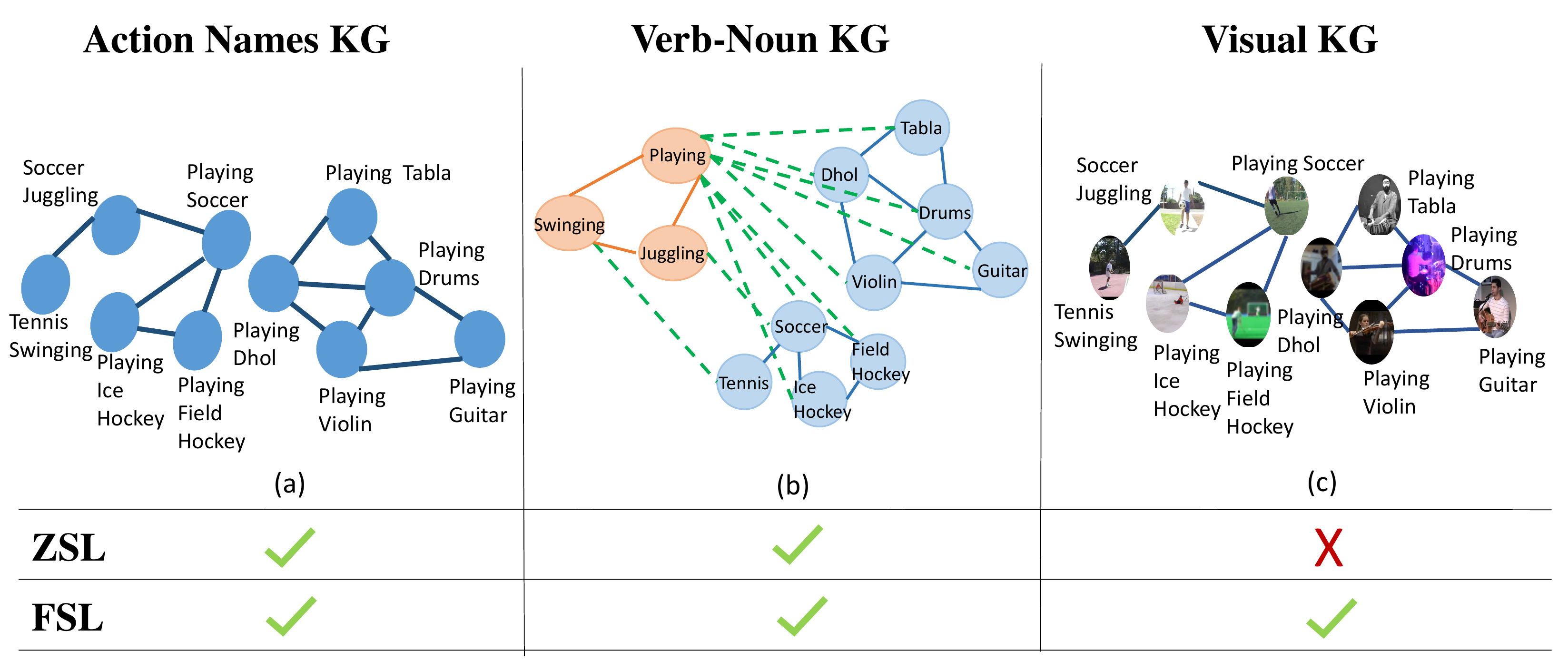}\\
\caption{We experiment with different Knowledge Graphs (KGs), using word and visual-feature based embeddings, for zero-shot learning and few-shot learning of actions. For zero-shot learning of actions, we construct a KG using action class names (i.e. \texttt{KG1}) (a) and a KG using the associated verb and nouns (i.e. \texttt{KG2}) (b). For few-shot learning, in addition, we use a KG with visual features (i.e. \texttt{KG3}) from a few examples from the test classes (c).}
\label{fig:teaser}
\end{figure*}%

Action recognition has seen rapid progress in the past few years, including better datasets~\cite{gu2018ava,kay2017kinetics} and stronger models~\cite{carreira2017quo,diba2017temporal,qiu2017learning,tran2018closer,wang2016temporal,xiang2018s3d,zhang2017deep}. Despite this progress, it is not easy to train an action classifier for a new category. A potential solution is to leverage the knowledge from seen or familiar categories to recognize unseen or unfamiliar categories. This is the zero-shot learning paradigm, where we transfer or adapt classifiers of related, known, or seen categories to classify unseen ones. Similarly, for few-shot action recognition, instead of testing on completely unseen classes, we have only a few labeled samples from the test classes, which help in learning about the rest of the test samples.

Both zero-shot and few-shot learning methods have been studied widely for image classification.  One of the recent technique involves building a knowledge graph (KG) representing relationships between seen and unseen classes and then training a graph convolutional network (GCN) on this KG to transfer classifier knowledge from seen to unseen classes~\cite{wang2018zero}. Using the same technique for action recognition is hard since, unlike objects, it is unclear what is the best knowledge representation for actions. One of the reasons as observed in~\cite{verb_noun} is that verbs have a broader definition and conflicting meaning.

In this work, we study the performance improvements by using different types of KGs for zero-shot and few-shot action recognition (Fig.\ref{fig:teaser}). The primary step in building a KG is generating a good implicit representation for action classes. In image classification, standard word embeddings (word2vec, GloVe, ConceptNet, etc.) capture the semantic knowledge associated with well-defined class names. However, for action classification, class names vary from single words (“sit”, “stand”, etc.) to phrases (“shooting ball (not playing baseball)”) and there are multiple definitions of the same (or similar) action class(es); like, “apply eye makeup” or “put on eye-liner”. 
Such diversity is less pronounced in image classification tasks due to the simplicity of labels. Our first contribution is studying different implicit representation for action classes and showing the advantages of a sentence2vector model in capturing the semantics of word sequences for zero/few-shot action recognition.

Our second contribution is building an explicit relationship map from these implicit representations of action classes. In image classification, the explicit representations for transferring knowledge from seen to unseen categories are using attributes or external KGs. Several datasets provide labeled class-attribute pairs (e.g., AwA~\cite{lampert2009learning} , aYahoo~\cite{farhadi2009describing}, COCO-Attributes~\cite{patterson2016coco}, MITstates~\cite{StatesAndTransformations}, etc.). Similarly, many KGs have nodes that correspond to image classification classes (e.g., WordNet~\cite{miller1995wordnet}, NELL, and NEIL~\cite{carlson2010toward,chen2013neil,wang2018zero}). In contrast, such sources are scarce for action classes. Wordnet contains verbs, therefore, it can be used to construct a KG for verbs, but we cannot have a KG with nodes representing the entire phrase (eg., ``playing(verb) guitar(noun)'') for an action class. Instead, there will be separate nodes for verbs and objects with defined inter-relationships. 
ConceptNet~\cite{conceptnet} has some phrases, but the list is not exhaustive and a lot of label names in our datasets are not present in ConceptNet.
On the other hand, we build a KG with an explicit relationship of the multi-word action phrases in any dataset. 
We append dataset with action classes from other datasets and construct two KGs, one for noun, and other for verb either by splitting the action phrase in cases like ``playing(verb) guitar(noun)" or using WordNet to get the nearest noun in cases like ``cake"(noun) for action class named ``baking"(verb). Further, we build a KG for few-shot learning using mean features of training data-points per class. We append this KG with the two KGs defined previously and observe performance improvement.

Finally, most previous work on zero-shot action recognition uses image-based learned models to estimate actions in videos. Recent advances in action recognition lead to the use of a network trained on video dataset as the feature extractor. So it requires an improved evaluation paradigm, since the action classes in the training set cannot be in the test set. We manually check for commonalities between the training datasets (Kinetics) and testing datasets (UCF101, HMDB51, Charades), but could not resolve problems within Kinetics which is a huge dataset and can have videos common across multiple classes. So we keep all Kinetics classes in training set and remove common classes from Kinetics with UCF101, HMDB51 and Charades from the test set. Hence, our third contribution is the creation of this evaluation paradigm using UCF101, HMDB51, Charades, and Kinetics datasets.

In summary our main three contributions are:
\begin{itemize}
    \item Better implicit representation of action phrases (which are word sequences) using sentence2vec
    \item Comparative study of different KGs for action zero-shot/few-shot learning
    \item Develop an improved evaluation paradigm for zero-shot/few-shot action recognition using networks trained on video datasets as feature extractors
\end{itemize}
These 3 contributions together builds an integrated approach for both zero-shot and few-shot learning.
\section{Related Work}
\label{rw}

\paragraph{\bf Action Recognition:}
Significant performance boost in state-of-the-art action recognition was observed with improved dense trajectories~\cite{wang2013action} and 3D ConvNets~\cite{ji20133d} which capture deep spatio-temporal features instead of handcrafted ones. Thereafter, multiple ideas like single stream networks~\cite{karpathy2014large}, two-stream networks~\cite{simonyan2014two}, end-to-end encoder-decoder based architectures~\cite{donahue2015long,tran2015learning,yao2015describing} and combining different streams with convolutional networks \cite{feichtenhofer2016convolutional,wang2016temporal} evolved. Recent studies include~\cite{carreira2017quo,diba2017temporal,qiu2017learning,tran2018closer,wang2016temporal,xiang2018s3d,zhang2017deep}. We use I3D model pre-trained on Kinetics described in~\cite{carreira2017quo}, to extract and learn features of the input videos.

\paragraph{\bf Zero-Shot Action Recognition:}

Zero-shot learning (ZSL) refers to the task of learning to predict on classes that are excluded from the training set~\cite{palatucci2009zero}. 
Various studies do ZSL for image classification and object detection~\cite{changpinyo2016synthesized,kodirov2015unsupervised,lampert2014attribute,sung2018learning}, and action recognition~\cite{alexiou2016exploring,gan2016recognizing,hahn2019action2vec,jain2015objects2action,mjain2,object_actor,xu2015semantic,xu2016multi,xu2017transductive,zhu2018towards}.
The other zero-shot action papers, to the best of our knowledge, mostly are not GCN based, which has been proven to do better than traditional zero-shot techniques for image classification~\cite{wang2018zero}. While ~\cite{gao2019know} is GCN based, their KG is very different from the one we use. They construct a single KG with actions and objects using ConceptNet~\cite{conceptnet}, where nodes are connected based on word embedding. They use visual object features as a second channel interconnected with the same edge weights to improve zero-shot learning. The number of objects in their graph is not dependent on the number of action classes. They show their best result when selecting 2000 most common visible objects in their dataset to get their object nodes, meaning they need access to the unlabelled test data (transductive). We use separate KGs for action, verb and noun and fuse them at the end with a fusion layer. Our verbs and nouns are dependent only on the action label and uses no visual information (inductive). 
We compare our results with ~\cite{gao2019know},~\cite{romera2015embarrassingly} and ~\cite{zhang2017learning}, where ~\cite{romera2015embarrassingly} uses a two linear layers network for learning relationships between features, attributes, and classes; while ~\cite{zhang2017learning} uses the image feature space to map the language embedding, instead of an intermediate space.

\paragraph{\bf Few-Shot Action Recognition:}
Few-shot for image classification has been explored using meta-learning for learning distance of samples and decision boundary in the embedding space~\cite{ml1,ml2,ml4}, or by learning the  optimization algorithm which can be generalized over different datasets~\cite{ml6,ml5}. A benchmark for few shot image classification is created in~\cite{fsl}. 

For action recognition, studies propose embedding a video as a matrix~\cite{fl_v2,fl_v1}, using deep networks~\cite{fl_v3} or generative models~\cite{fl_v5,fl_v3} and using human-object interaction~\cite{comp_lr}. We tried GCN based few-shot learning for action recognition, but our approach cannot be compared to many of these approaches due to two reasons -- 1) Each paper uses a different dataset split, and our splits are different as well because we use a pre-trained network from Kinetics in our pipeline; 2) We do not evaluate the episodic learning formulation like several other papers. Our aim is to improve few-shot using the KG constructed for the zero-shot setting (relationship of class names, etc.) thereby building an unified zero-shot and few-shot learning framework, which to the best of our knowledge, is not explored in the past.

\begin{figure*}
\centering
\includegraphics[width=\linewidth]{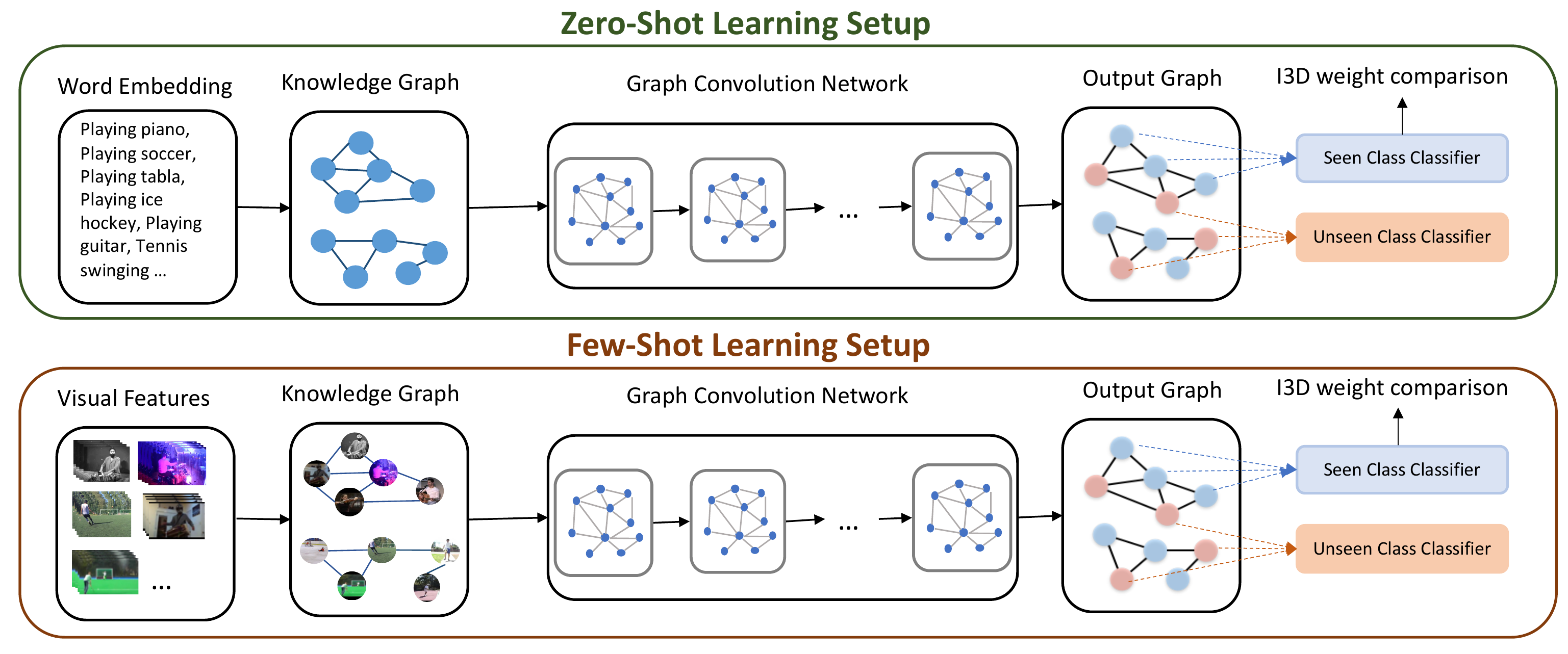}
\caption{System overview: We use knowledge graphs based on word embeddings (action class names, and associated verbs and nouns) and visual features for action recognition. With the word embeddings based knowledge graph, we propose a zero-shot learning approach and with visual features based knowledge graph we propose a few-shot learning approach.}
\label{fig:system}

\end{figure*}

\paragraph{\bf Knowledge Graphs and Graph Convolution Networks:}



KGs are used to improve performance for different visual applications~\cite{marino2016more,fang2017object}. Automatic construction of a large KG and relationship learning has captured a lot of attention in the past~\cite{bordes2014constructing,choudhury2017nous,gao2019know,lin2015learning}. 

We focus on construction of a KG to depict inter-relationships of action categories. \cite{verb_noun} shows how verbs and nouns have different levels of complexities and usually an action phrase comprises of both or just the verb. We explore different KGs, including one with verbs and nouns only, to understand how these knowledge graphs improve performance for action recognition in zero-shot and few-shot learning setup.

To process graphs using deep learning algorithms, graph convolution networks(GCN) have been used for a number of different applications including action recognition~\cite{yan2018spatial,ghosh2020stacked}. 
For GCNs, some of the initial works include~\cite{atwood2016diffusion,duvenaud2015convolutional,henaff2015deep}. Concept of spectral graph theory was introduced in ~\cite{hammond2011wavelets} and extended in ~\cite{defferrard2016convolutional}, to a formal version of GCN. \cite{kipf2016semi} simplified GCNs through localized first order approximations of spectral GCN. We use the GCN version developed by~\cite{kipf2016semi}.
Graph attention networks~\cite{velivckovic2017graph}, GraphFlow~\cite{ji2019graphflow} and Long Tail relation extraction~\cite{zhang2019long} are some of the recent developments in this field.

~\cite{wang2018zero} uses Graph Convolution network on KG for zero-shot image classification. The knowledge graph was formed with NELL(Never Ending Language Learning)~\cite{carlson2010toward}, NEIL(Never Ending Image Learning)~\cite{chen2013neil} and WordNet~\cite{miller1995wordnet}. We use a similar model based on GCN for actions.

\section{Overview of GCN}
\label{sec:gcn_o}

The implementation technique of Graph Convolutional Networks in \cite{kipf2016semi} is used to train our KG to transfer classifier layer weights from trained classes to unseen test classes. The GCN operation can be described by the equation
$$
H^{l+1} = g(H^{l},A) = \sigma(\hat{D}^{-1/2}\hat{A}\hat{D}^{-1/2}H^{l}W^{l})
$$
where $\hat{A}=I+A$ and ${A}$ is the adjacency matrix consisting of edge weights between nodes, $\hat{D}$ is the node degree matrix of $\hat{A}$, $H^l$ and ${W^l}$ are the $N\times d^{l}$ input matrix of the $l^\text{th}$ layer and $d^{l} \times d^{l+1}$ weight matrix respectively. $N$ is the number of nodes in the graph, $d^{l}$ is the dimension of the $l^\text{th}$ layer and $\sigma$ represents a non-linear activation function (e.g., ReLU).
 
Zero-shot/few-shot action recognition using GCN (Fig.\ref{fig:system}) follows a similar technique as \cite{wang2018zero}. It consists of training and testing phases as described next.

\paragraph{\bf Training:}
Initially, a model pre-trained on Kinetics is fine-tuned using training classes of UCF101, HMDB51, or Charades, followed by the extraction of the final classifier layer weights to be used for training the GCN.
The constructed KG, along with the adjacency matrix, are inputs to the GCN. The output of each node of the GCN has the same dimensions as the trained classifier layer filter size (1024 in our case). The GCN is trained such that its output for the training classes matches the classifier layer weights of the trained I3D model. The loss used is the mean squared error (MSE) loss. 

So if there are $C_\text{train}$ number of training classes, $C_\text{test}$ number of test classes and the output feature dimension of each class is $d$, then the output of the GCN, $W_\text{GCN}$, is of size $(C_\text{train}+C_\text{test}) \times d$. From $W_\text{GCN}$, the output dimensions corresponding to the training nodes are selected, denoted by $W_\text{GCNTrain}$ with size $C_\text{train} \times d$. This feature is of the same dimension as the weights of the I3D classifier layer trained or fine-tuned on the training classes of the dataset, $W_\text{cls}$. The MSE loss that is back-propagated is given by $\lVert W_\text{GCNTrain}-W_\text{cls} \rVert_{2}$.

\paragraph{\bf Testing:} 
During test time, the penultimate layer of the I3D model is used to extract the features of the test images $f_\text{test}$ with dimensions $N\times d$. The output of the test nodes of the GCN with dimension $C_\text{test} \times d$ is extracted from $W_\text{GCN}$, denoted by by $W_\text{GCNTest}$. The output class probabilities for the test images ($P_\text{test}$) are obtained as $P_\text{test} = f_\text{test}W_\text{GCNTest}^{T}$.

\section{Proposed Knowledge Graphs for Actions}
\label{sec:akg}

In this section, we describe the construction of different KG for actions. We follow similar pipeline as \cite{wang2018zero} (also described in Section\ref{sec:gcn_o}) which requires a KG as input. \cite{wang2018zero} use Wordnet embeddings to construct the KG for ZSL on image classification. Compared to~\cite{wang2018zero}, our action label classes are sentences or phrases instead of words, which is why using wordnet or word2vec doesn't provide distributive and coherent embeddings for action labels. Moreover, getting semantically correlated embedding space for words and visual features for a good KG is another challenge. We describe these challenges and how we tackle them while constructing three different versions of KGs for actions.

\paragraph{\texttt{\bf {KG1:}}} The first KG is based on word descriptors of action class names. Since our action classes are composed of multiple words like a sentence or phrase, averaging word2vec embedding for all words in the sentence does not provide a cohesive embedding space. We discuss the experimental results for word2vec embeddings in Section\ref{sec:disc}. To overcome this challenge, we use the sentence2vec model described in \cite{pagliardini2017unsupervised}, which is an unsupervised learning method to learn embeddings for whole sentences. We use the unigrams model trained on Wikipedia to generate our sentence embeddings. 

The node features in \texttt{KG1} are the sentence embeddings. The nodes from Kinetics action classes are added in \texttt{KG1} corresponding to each dataset (UCF101, HMDB51, and Charades). This is inspired by \cite{xu2016multi,xu2015semantic}, where they show distinct advantages of adding classes and images from other datasets in zero-shot learning. Although we cannot directly add images due to the way our model is constructed, we add new activity classes from the Kinetics dataset to increase the size of our KGs. Appending 400 Kinetics classes to UCF101 results in a total of 501 nodes in the \texttt{KG1} for UCF. Similarly appending the nodes to HBDB51 and Charades results in a total of 451 nodes and 557 nodes respectively. We show more results on performance comparison with and without adding Kinetics nodes in Section~\ref{sec:disc}.

With the sentence2vector node features, we construct the \texttt{KG1} where node ${i}$ is connected to another node ${j}$ in the combined dataset based on edge weights $A_{ij}$ from cosine similarity of node features. Here, $A$ is the adjacency matrix for \texttt{KG1}. We sort the edges weights in descending order to get the top N closest neighbors per node. N is a hyperparameter that is determined experimentally and is dependent on the dataset. It is 5 for HMDB51 and UCF101 and 20 for Charades. ${j}$ being one of the top N neighbors of ${i}$ does not mean that the vice versa is true as well. To make the adjacency matrix symmetric, we fill $A_{ji}$ with the same value as $A_{ij}$, so the number of connections to each node $>=N$.

\paragraph{\texttt{\bf {KG2:}}} The second graph, \texttt{KG2}, is constructed with verbs and nouns associated with each action class. This graph is inspired by multiple works on zero-shot action using human object interaction where the detected objects in the scene are used to draw the relationships between seen and unseen action classes~\cite{gao2019know,jain2015objects2action}. In ~\cite{gao2019know} object detection is carried out in the visual domain as well and then mapped to word domain for zero-shot learning. We do not do mapping for objects features from visual to word. Instead, we just take the output of verb and noun graphs (\texttt{KG2}), and pass it through the fusion layer to get the visual action (noun+verb) classifier weights. 

To construct \texttt{KG2}, we use a standard language lemmatizer~\cite{nltk} to break up a phrase describing an action and convert the word to its root form. Then, we use a part-of-speech (pos)~\cite{pos} tagger to label the word as a noun or a verb. Still, a lot of action class names do not have a noun in the phrase, for example ``beatboxing''. For such classes the pos tagger gives a noun label of ``unknown'' and if Wordnet can return a noun that is related to that word, we replace the ``unknown'' by the noun. 
For action classes like ``archery'', which does not have a specific verb associated with it, we replace the verb with ``doing''. For node features, we compute sentence2vec embeddings as above for verbs and nouns. Hence, we get a set of graphs with only verbs and only nouns. These also have same number of nodes as \texttt{KG1}. Moreover, these graphs are used and categorized together as \texttt{KG2}, since they provide partial information about action class (either verb or noun). \texttt{KG1} and \texttt{KG2} can be used to define ZSL setup.

\paragraph{\texttt{\bf {KG3:}}} The third graph is developed to see relative performance improvements by incorporating only a few labelled images per test class. We use averaged visual features as nodes in \texttt{KG3}. In the visual feature space, we see implicit clustering of similar actions, which is sometimes not captured in word embedding space. For example, ``pommel horse'' and ``horse walking'' are considered similar in word embedding space, but these are very different activities which is captured in visual embedding space shown for dataset UCF101 in Fig.~\ref{fig:tsne_feat}. We randomly pick 5 videos from each test class and use I3D to generate video features as described in Section~\ref{feat_extract}. Then taking the mean of these features, we get the graph node descriptors and take their cosine similarity to generate the adjacency matrix as we do for \texttt{KG1} and \texttt{KG2}. This generates a graph based on visual features. \texttt{KG3} is used to replicate few-shot learning setup using KGs, since we use 5 visual samples for each test class to construct the nodes. In few-shot setting, we can combine \texttt{KG3} with \texttt{KG1} and \texttt{KG2} to improve results.

\section{Experimental Setup}

\subsection{Datasets}

We use following four datasets, where Kinetics is just for pre-training the model, and rest are used for experiments:

\paragraph{\bf{Kinetics}~\cite{kay2017kinetics}:} Kinetics is a large dataset with 400 classes and about $3*10^{6}$ videos. We do not actually need access to Kinetics videos, but the class names and an I3D model pre-trained on Kinetics available in \cite{carreira2017quo}. Since we use Kinetics for pre-training I3D and data augmentation while training the GCN, we cannot keep common classes between Kinetics and UCF101 or HMDB51 or Charades in the test set while doing zero-shot learning. So, we use classes in UCF101, HMDB51 and Charades that are also present in Kinetics, as training set.

\paragraph{\bf{UCF101}~\cite{soomro2012ucf101}:} UCF101 has 13320 videos from 101 classes. After removing common classes with Kinetics, we get 23 classes with 3004 videos in test set for UCF101 and the remaining 78 classes are used for training. Some test class labels do not have semantically correlated neighbors. So, we appended these class names with extra words, for example ``front crawl" in UCF101 becomes ``front crawl swimming". We discuss class-wise accuracy for test classes in Fig.\ref{fig:bar_plot}.

\paragraph{\bf{HMDB51}~\cite{kuehne2013hmdb51}:} HMDB51 has 6849 videos from 51 classes. Similar to UCF101, we remove common classes with Kinetics, and get 12 classes with 1541 videos for HMDB51's test set and remaining 39 classes for training. Additionally, to encourage correlation with action classes in Kinetics, we convert the class labels to continuous tenses. For example, classes like ``eat", uses sentence2vector embedding corresponding to ``eating".

\paragraph{\bf{Charades}~\cite{charades2016}:} Charades has 9848 videos from 157 classes and is also a multilabel dataset, meaning each video can have multiple action labels. Charades has noun and verb labels associated with each action class, which we use directly without labelling ourselves. After removing all videos which have at least one common label with Kinetics, we are left with 111 possible test classes. Each video can have both training and test labels in Charades. We cannot separate the training and test videos but just the classes. We split the classes into 50-50 train-test split meaning there are 79 and 78 train and test classes respectively. The 78 test classes are from the 111 classes not in common with Kinetics. All videos with at least one training class are kept in training set and we remove test class labels from them. The rest of the videos are test videos and training class labels are removed from them.

\begin{figure}
\centering
 \includegraphics[width=\linewidth]{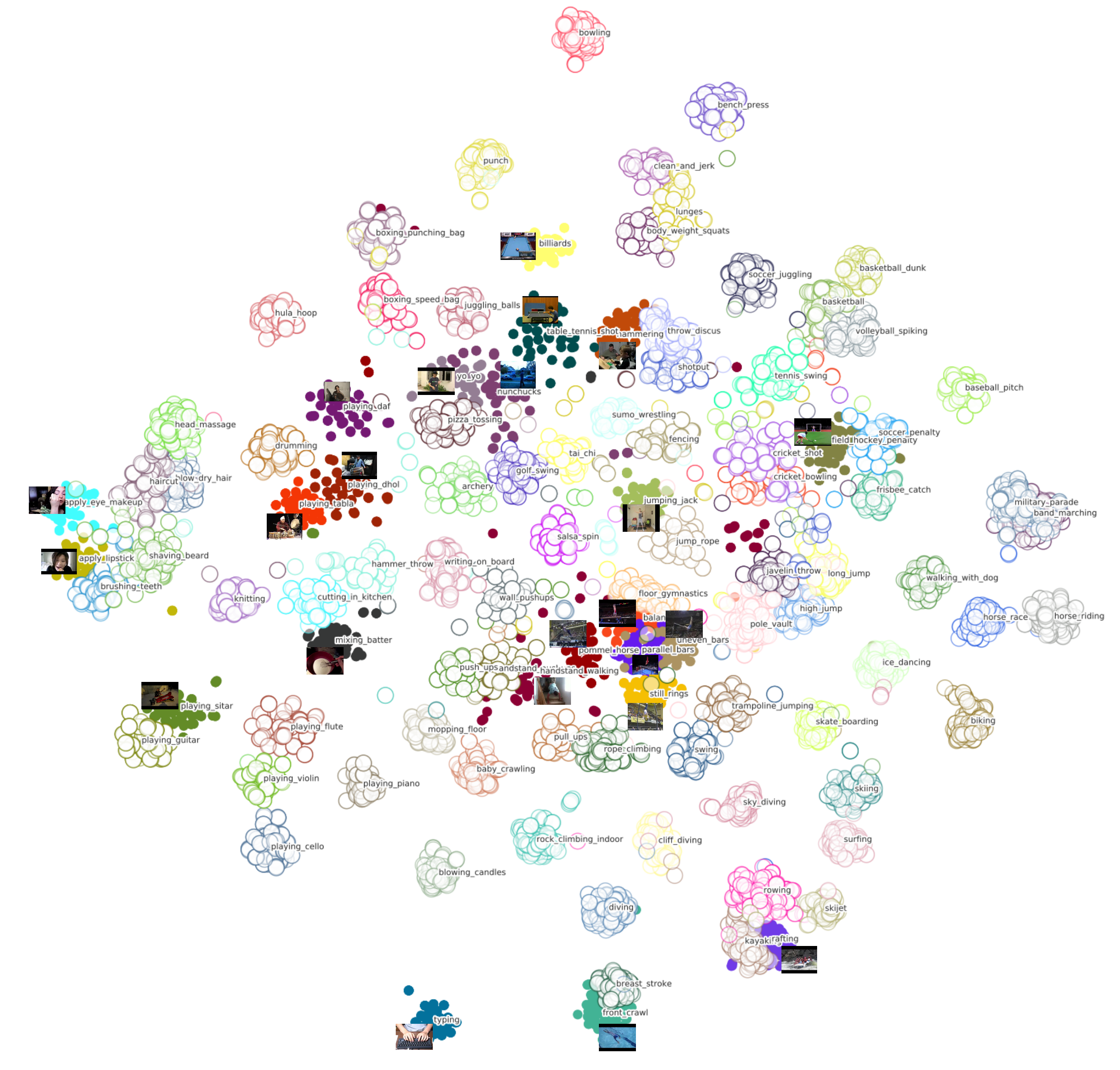}
\caption{t-SNE visualization showing feature distribution of UCF101 video dataset. Sample images are added for our test classes. (Best viewed in digital format)}
\label{fig:tsne_feat}
\end{figure}

\subsection{Feature Extraction} 

\label{feat_extract} 

To extract \emph{video features}, we use initial model of I3D trained on Kinetics data and fine-tune the last layer on the training classes of either UCF101 or HMDB51. For Charades, just fine-tuning the last layer did not yield good classification performance, so we fine-tune the whole network. This means while training, we cannot compute loss on the Kinetics nodes in the KG for Charades. Even after fine-tuning the complete network for Charades we did not achieve significant performance for zero-shot learning; so we use inverse cross-correlation of training features multiplied with itself as last layer weight inspired by~\cite{romera2015embarrassingly}, to train GCN. We visualize the video feature space distribution of the UCF101 classes in Fig.\ref{fig:tsne_feat} with some example images for the test classes. As we can see in Fig.\ref{fig:tsne_feat}, similar classes are grouped together forming clusters.

\begin{table}
\centering
\caption{Zero-shot learning results for all 3 datasets where we compare performances of \texttt{KG1}, \texttt{KG2} and a combination of the two. \texttt{KG1}+\texttt{KG2} always does the best. For UCF101 and HMDB51, the results are in mean accuracy whereas for Charades, we report mean average precision (mAP)}
\resizebox{0.7\linewidth}{!}{
\begin{tabular}{@{}lccc@{}}
\toprule
Dataset & \texttt{KG1} & \texttt{KG2} & \texttt{KG1}+\texttt{KG2} \\
\midrule
UCF101 & 49.14 & 45.47 & \bf{50.13}\\
HMDB51 & 38.01 & 31.57 & \bf{40.77} \\
Charades & 15.81 & 12.48 & \bf{18.21}\\
\bottomrule
\end{tabular}}
\label{table:results2}
\end{table}

\begin{table*}
\centering
\caption{Zero-shot learning results for all 3 datasets. The baselines are ESZSL, DEM, Objects2Action, UR, CEWGAN and TS-GCN. For UCF101 and HMDB51, the results are in mean accuracy whereas for Charades, we report mean average precision (mAP) since it is multi-label dataset.}
\resizebox{0.95\linewidth}{!}{
\begin{tabular}{@{}lccccc@{}}
\toprule
 Method & \multicolumn{2}{c}{UCF101} & \multicolumn{2}{c}{HMDB51} & Charades \\
 & 23-78 split & 50-51 split & 12-39 split & 25-26 split & 78-79 split \\
\midrule
ESZSL~\cite{romera2015embarrassingly} & 35.27 & 15.0 & 34.16 & 18.5 & 17.21 \\
DEM~\cite{zhang2017learning} & 34.26 & - & 35.26 & - & - \\
Objects2Action~\cite{jain2015objects2action} & - & 30.3 & - & 15.6 & - \\
UR~\cite{zhu2018towards} & - & 17.5 & - & 24.4 & - \\
CEWGAN~\cite{mandal2019out} & - & 26.9 & - & 30.2 & - \\
TS-GCN~\cite{gao2019know} & 44.5 & 34.2 & - & 23.2 & -\\
\bf{Ours} & \bf{50.13} & - & \bf{40.19} & - &  \bf{18.21} \\
\bottomrule
\end{tabular}}
\label{table:results_STA}
\end{table*}

\subsection{Our Pipeline}
Our GCN consists of 6 layers with intermediate layer filter dimensions of $512 \rightarrow 1024 \rightarrow 1024 \rightarrow 1024 \rightarrow 1024 $. We choose 6 layers empirically. Our hypothesis is that lower depth might reduce field of view, necessary for information transfer, whereas higher depths might result in over-smoothing. The convolution kernel is of size 1. For training/fine-tuning both I3D model and the GCN model, we use ADAM optimizer with initial learning rate of 0.001. A stepwise scheduler with a drop rate of 0.99 after every 100 epochs is used for I3D training. For GCN, stepwise scheduler drop rate is 0.999 after every 100 epochs. Classwise mean accuracy is used as the evaluation metric for UCF101 and HMDB51 and mean average precision (mAP) scores for Charades. Most of the training parameters are the same for few-shot learning setup as well, except we use a smaller learning rate of 0.00005 for UCF101.

To fuse the outputs of the different KGs, we concatenate along the channel dimension and then pass them through a GCN layer. For zero-shot this fusion GCN layer uses Adjacency matrix of \texttt{KG1} and for few-shot it uses the adjacency matrix of \texttt{KG3}. For \texttt{KG1}+\texttt{KG2} in UCF101, the above fusion technique did not give good performance. So, we use the weighted sum of the outputs of \texttt{KG1} and \texttt{KG2} with weights of 0.9 for \texttt{KG1} and 0.05 each for the verb and noun from \texttt{KG2}.

\section{Results}
\label{results}

The results for zero-shot learning on all 23 test classes for UCF101, 12 test classes for HMDB51 and 78 test classes for Charades are in Table\ref{table:results2}. These results are based on KGs \texttt{KG1} and \texttt{KG2} and combination of both. The combination of \texttt{KG1} and \texttt{KG2} graph is done by passing it through the fusion layer (for HMDB51 and Charades) or weighted summation of output (for UCF101). Since all datasets have many action classes without any nouns, only \texttt{KG2} did not give good performance, but the combination of \texttt{KG1}+\texttt{KG2} works well.

We also provide the comparison with state-of-the-art in Table~\ref{table:results_STA}. For our data split, we have compared our results with three previous works carried out under similar zero-shot learning settings,  ESZSL~\cite{romera2015embarrassingly}, DEM~\cite{zhang2017learning} and TS-GCN~\cite{gao2019know}. We could not apply DEM baseline results for Charades, since it is a multi-label dataset. Also, TS-GCN only released code for the transductive setup for UCF101. We have implemented the inductive version and compared to it. We have also added some of the recent results for zero-shot learning. Either their splits are different, or they do not provide code, or an essential part of their framework is missing. However, note that recent work of \cite{gao2019know} outperforms these other approaches on their splits and we outperform \cite{gao2019know} on our splits.

We report results for combining \texttt{KG3} with \texttt{KG1} and \texttt{KG2} in Table~\ref{table:results3}. Since we are using \texttt{KG3}, these experiments can be considered as few shot learning setup. To create a baseline, we used the nearest neighbor search to get the class label for test videos. Based on the 5 labelled videos provided, we calculate the mean or center feature for each class and then we use cosine distances between the rest of the test videos and these class centers to sort them into corresponding classes.
Our results along with the baselines are in Table~\ref{table:results3}. We use the same train-test splits for UCF101 and HMDB51. For both UCF101 and HMDB51, we get best results if we use all 3 KGs. We do not conduct this experiment for Charades since each video has multiple labels, hence each video data point will update multiple class centers resulting in overlapping class distribution.


\begin{table*}
\centering
\caption{Few-shot learning results for the UCF101 and HMDB51 datasets. The baseline is nearest neighbor, given 5 videos for each test set. The combination of \texttt{KG1}, \texttt{KG2} and \texttt{KG3} does the best in both cases.}
\resizebox{0.6\linewidth}{!}{
\begin{tabular}{@{}lccccc@{}}
\toprule
Dataset & Baseline & \texttt{KG3} & \texttt{KG3}+\texttt{KG1} & \texttt{K3}+\texttt{KG2} & \texttt{KG3}+\texttt{KG1}+\texttt{KG2}\\
\midrule
UCF101 & 52.7 & 57.04 & 62.10 & 59.92 & \bf{64.24}\\
HMDB & 30.2 & 45.07 & 45.67 & 47.61 & \bf{47.69}\\
\bottomrule
\end{tabular}}
\label{table:results3}
\end{table*}
\section{Analysis}
\label{sec:disc}

\paragraph{\bf Word embeddings for action labels:}

\begin{figure}
\centering
  \includegraphics[width=\linewidth]{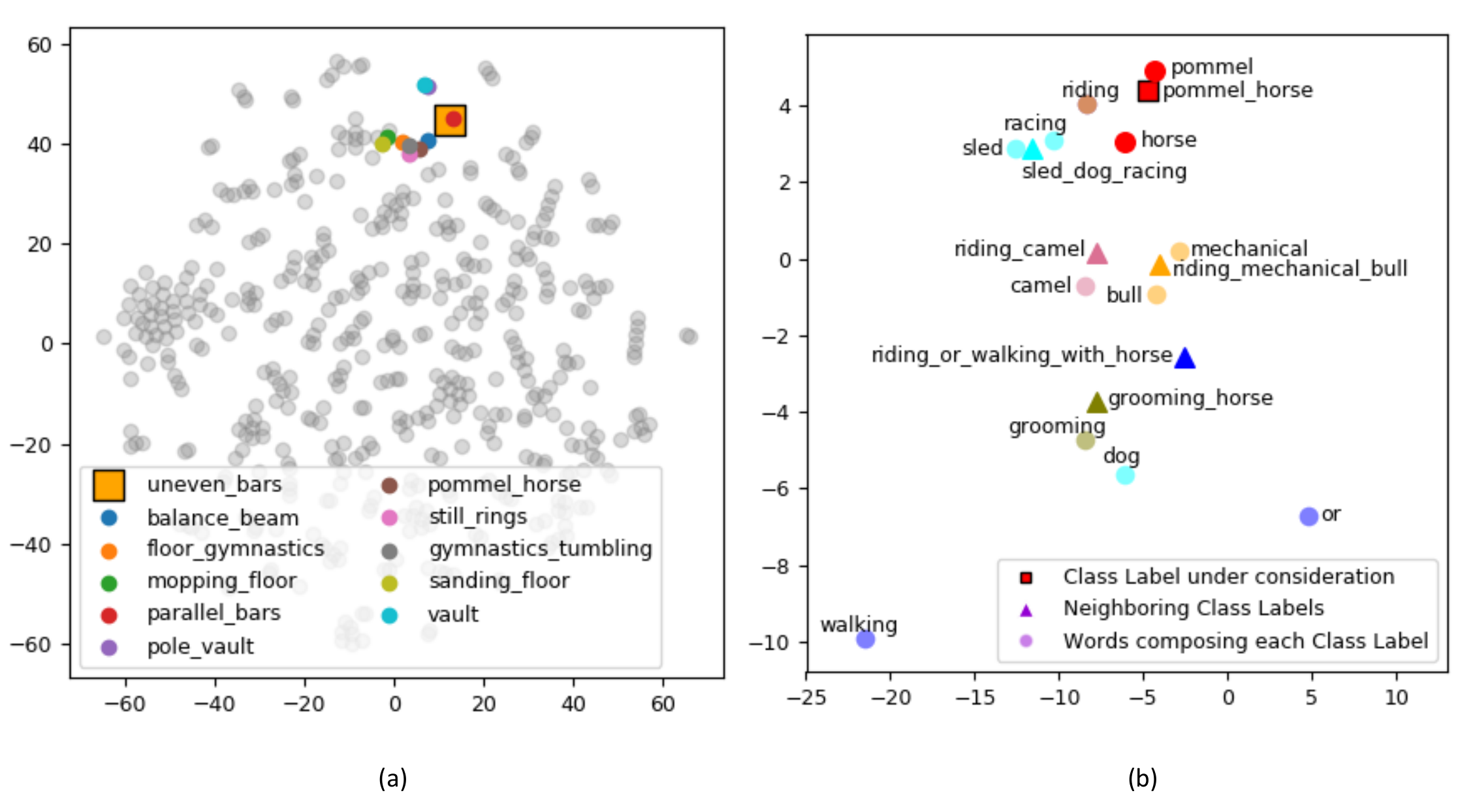}
  \caption{(a) Sentence2Vec embedding space for Kinetics and UCF101 classes. The class ``uneven bars'' and its neighbors are highlighted. (b) Class ``Pommel horse'' and its neighboring classes in Kinetics dataset using word2vec embedding. The embeddings of each individual word forming the phrase is also displayed. (Best viewed in digital format)}
  \label{fig:s2v}
\end{figure}

For constructing node features from action labels, we used the word2-vec embeddings trained on Google News~\cite{mikolov2013efficient,mikolov2013distributed,mikolov2013linguistic}. For all words in each class name, the word2vec embeddings were averaged to give a resultant embedding for the whole phrase, which serves as features of the nodes in the KG. In Fig.\ref{fig:s2v}(b), we show the word2vec embedding space of node ``Pommel Horse" and its nearest neighbor class nodes.

Averaging word2Vec embedding for all words in action class label phrase works in some cases, but it cannot always capture the meaning or correct relationships between the action classes. Hence, for a class like ``riding or walking with horse'' in Kinetics dataset, the embedding for each word is located far apart from each other as displayed in Fig.\ref{fig:s2v}(b). The mean of these individual words does not lie close to related words in the embedding space and hence does not capture meaningful information.

\begin{table}
\centering
\caption{Performance comparison between word2vec embedding and sentence2vec embedding based models. Both the models are trained on graphs consisting of class nodes from Kinetics and UCF101 with losses on both. Performance metric used is mean accuracy.}
\resizebox{0.6\linewidth}{!}{
\begin{tabular}{@{}lc@{}}
\toprule
Method & Mean Accuracy\\
\midrule
Word2Vec & 38.02  \\
Sentence2vec & 49.14 \\
\bottomrule
\end{tabular}}
\label{table:w2v_s2v}
\end{table}

To solve this problem we use sentence2vec model from \cite{pagliardini2017unsupervised}, which captures the semantic meaning of sequences of words. Using this embedding space, the closest word match to a class like ``uneven bars" is ``gymnastics tumbling". The word embedding space for all the classes in UCF101 and Kinetics are displayed in Fig.\ref{fig:s2v}(a).
The word ``Uneven bars'' along with its neighbors are emphasized.
We run experiments with both word2vec embeddings trained on Google News~\cite{mikolov2013efficient,mikolov2013distributed,mikolov2013linguistic} 
and Sentence2Vec embeddings based on unigram model trained on Wikipedia \cite{pagliardini2017unsupervised}. 
The results on UCF101 are shown in Table \ref{table:w2v_s2v}. These results show significant improvement by using setence2vec over word2vec for \texttt{KG1}.


\begin{table}
\centering
\caption{Experiments with 3 different knowledge graph constructions. The variations are due to using only UCF101/HMDB51 classes for the knowledge graph or appending it with Kinetics classes and training loss being calculated on UCF101/HMDB51 nodes only or both UCF101/HMDB51 and Kinetics nodes in the knowledge graphs. Performance metric used is mean accuracy.}
\begin{tabular}{@{}lcc@{}}
\toprule
Knowledge & Nodes for Loss & Mean \\
Graph & Computation & Accuracy\\
\midrule
UCF only & UCF   & 27.72  \\
UCF+Kinetics & UCF  & 32.85 \\
\bf{UCF+Kinetics} & \bf{UCF+Kinetics} & \bf{49.14} \\
\midrule
HMDB only & HMDB   & 31.09  \\
HMDB+Kinetics & HMDB & 29.22 \\
\bf{HMDB+Kinetics} & \bf{HMDB+Kinetics} & \bf{38.01} \\
\bottomrule
\end{tabular}
\label{table:graph_diff_abl}
\end{table}

\paragraph{\bf Appending Knowledge Graphs with more action classes:}
We augment the UCF101 and HMDB51 KGs with Kinetics class labels in three different ways. 
In the first configuration, either only the UCF101 nodes or HMDB51 nodes are used in the KG (101/51 nodes) out of which, 78 and 39 are training nodes respectively. The loss is computed by comparing the output of the GCN on these classes to the weights in the final classifier layer of the fine-tuned I3D network.


The second configuration uses the same KG as \texttt{KG1} explained in section~\ref{sec:akg}. The loss is computed by comparing the output of only the UCF101 or HMDB51 training nodes (78/39 nodes) to the final classifier layer of the fine-tuned I3D network.

In the third configuration, again \texttt{KG1} is used. Although, now the loss is computed by summing the 2 MSE losses:
(a) Loss 1 by comparing the output of only the UCF101 or HMDB training nodes(78/39 nodes) to the final classifier layer of the fine-tuned I3D network.
(b) Loss 2 by comparing the output of the Kinetics nodes (400 nodes) to the classifier layer weight of I3D pre-trained on Kinetics.
The results of these three experiments are shown in Table \ref{table:graph_diff_abl}. For UCF101 and HMDB51, third configuration works best. 



\begin{figure*}
\centering
 \includegraphics[width=\linewidth]{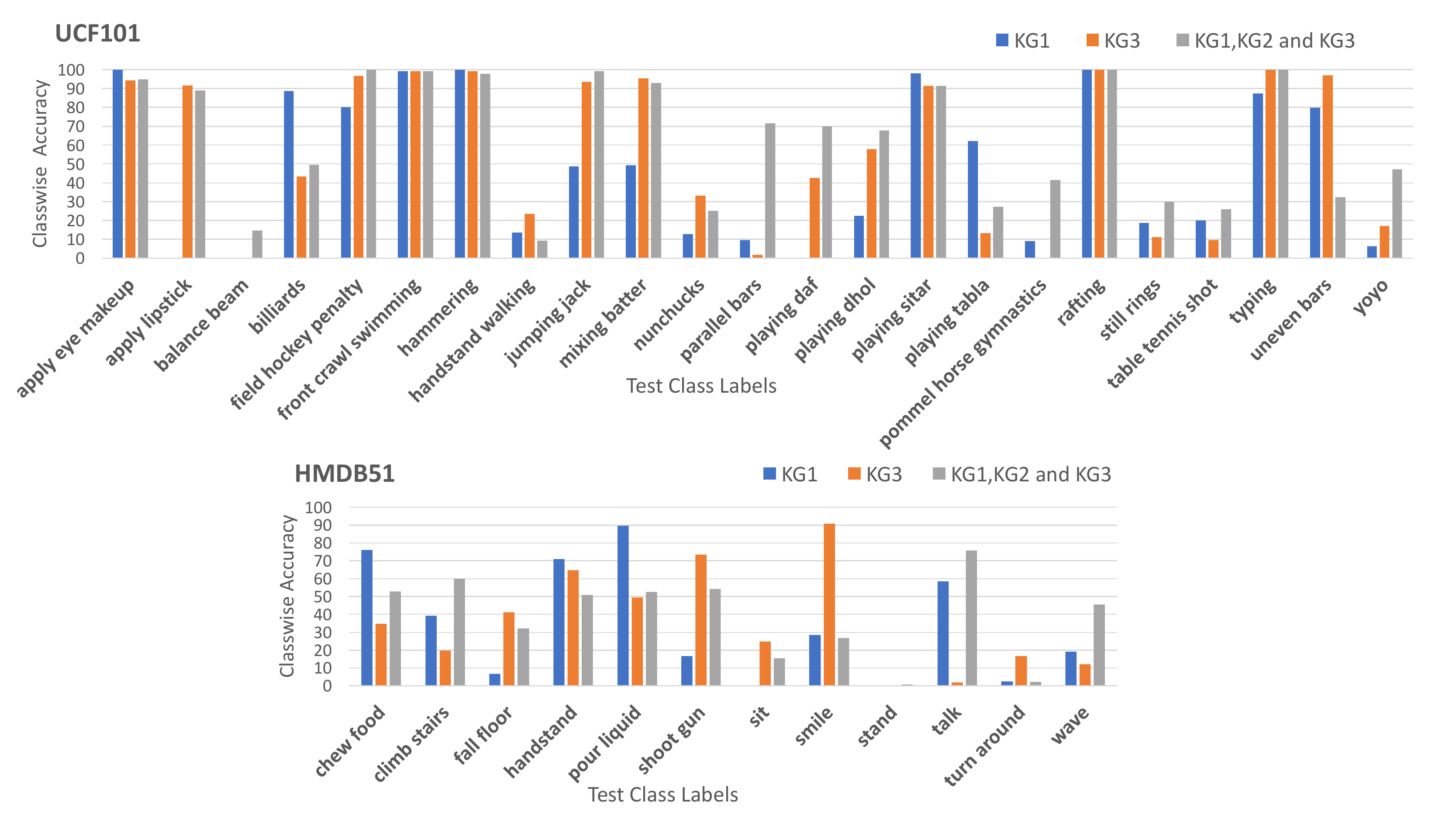}
\caption{This figure shows class-wise accuracy for different KGs and combination of KGs for UCF101 and HMDB51. We added few words for better word embeddings in the labels (such as ``pommel horse'' becomes ``pommel horse gymnastics''), which improves performance for only word based KG (i.e. \texttt{KG1}), as shown here. Each color for bar represents a KG, blue is word based KG, orange is visual feature based KG and grey is combination of all three KGs (\texttt{KG1},\texttt{KG2} and \texttt{KG3}).}
\label{fig:bar_plot}
\end{figure*}

\paragraph{\bf Types of connections in Knowledge Graphs:}

While constructing the KG with both UCF101 or HMDB51 and Kinetics dataset, we used two types of graph connections. In fully-connected graphs all nodes can be connected to all other nodes, out of which we select top 5 connections. In bipartite, for every node in UCF101 or HMDB51 dataset, we find the top 5 connections to the Kinetics dataset nodes and vice versa. The fully connected(FC) graph works better than the bipartite graph (Table \ref{table:fully_vs_bipartite}).

\begin{table}
\centering
\caption{Performance comparison for fully connected(FC) and bipartite graphs constructed with UCF101 or HMDB51 with Kinetics dataset nodes. Both the models are trained on graphs consisting of class nodes from two datasets (UCF101 and Kinetics or HMDB51 and Kinetics) with losses on both. Performance metric used is mean accuracy.}
\begin{tabular}{@{}lcc@{}}
\toprule
Method & Mean-accuracy & Mean-accuracy\\
& for UCF & for HMDB \\ 
\midrule
FC & 49.14 & 38.01\\
Bipartite & 33.11 & 28.49\\
\bottomrule
\end{tabular}
\label{table:fully_vs_bipartite}
\end{table}


\paragraph{\bf Analysis of Class-wise Accuracy using different Knowledge Graphs:}

\begin{table}
  \centering
  \caption{Performance comparison of using GCN vs a linear combination (using the adjacency matrix edge weights) of the top 4 closest training class weights to the test classes. Performance metric used is mean accuracy.}
  \begin{tabular}{@{}lc@{}}
    \toprule
    Method & Mean accuracy\\
    \midrule
    GCN & 49.14  \\
    Linear Combination & 42.57 \\
    \bottomrule
    \end{tabular}
    \label{table:linear_combo}
\end{table}%

\begin{table}
  \centering
  \caption{Performance comparison of using an encoder-decode layer before the GCN layers on UCF101 dataset vs not using one. Performance metric used is mean accuracy.}
    \begin{tabular}{@{}lc@{}}
    \toprule
    Method & Mean accuracy\\
    \midrule
    without encoder-decoder & 49.14  \\
    with encoder-decoder & 47.72 \\
    \bottomrule
    \end{tabular}
    \label{table:enco_deco}
\end{table}

\begin{table*}
\centering
\caption{Results on UCF101 with 10 randomly selected test classes leaving 91 classes to be used for training I3D and GCN. Mean accuracy is used for evaluation. The experiments are carried out 5 times and the final column provides the mean accuracy scores. We compare our results to two previous work with similar settings.}
\resizebox{0.85\linewidth}{!}{
\begin{tabular}{@{}llcccccc@{}}
\toprule
Method & Nodes for Loss Computation & Split 1 & Split 2 & Split 3 & Split 4 & Split 5 & \bf{Mean}\\
\midrule
ESZSL & - & 61.25 & 60.30 & 53.68 & 64.81 & 60.56 & 60.12 \\
DEM & - & 60.87 & 65.88 & 41.89 & 61.90 & 52.11 & 56.53 \\
\bf{Ours} & UCF101 & 59.68 & 48.51 & 42.18 & 49.86 & 43.12 & 48.67 \\
\bf{Ours} & UCF101+Kinetics &
\bf{83.62} & \bf{72.60} & \bf{71.57} & \bf{70.85} & \bf{49.39} & \bf{69.61} \\
\bottomrule
\end{tabular}
}
\label{table:results1}
\end{table*}

\begin{figure}
\begin{center}
\includegraphics[width=\linewidth]{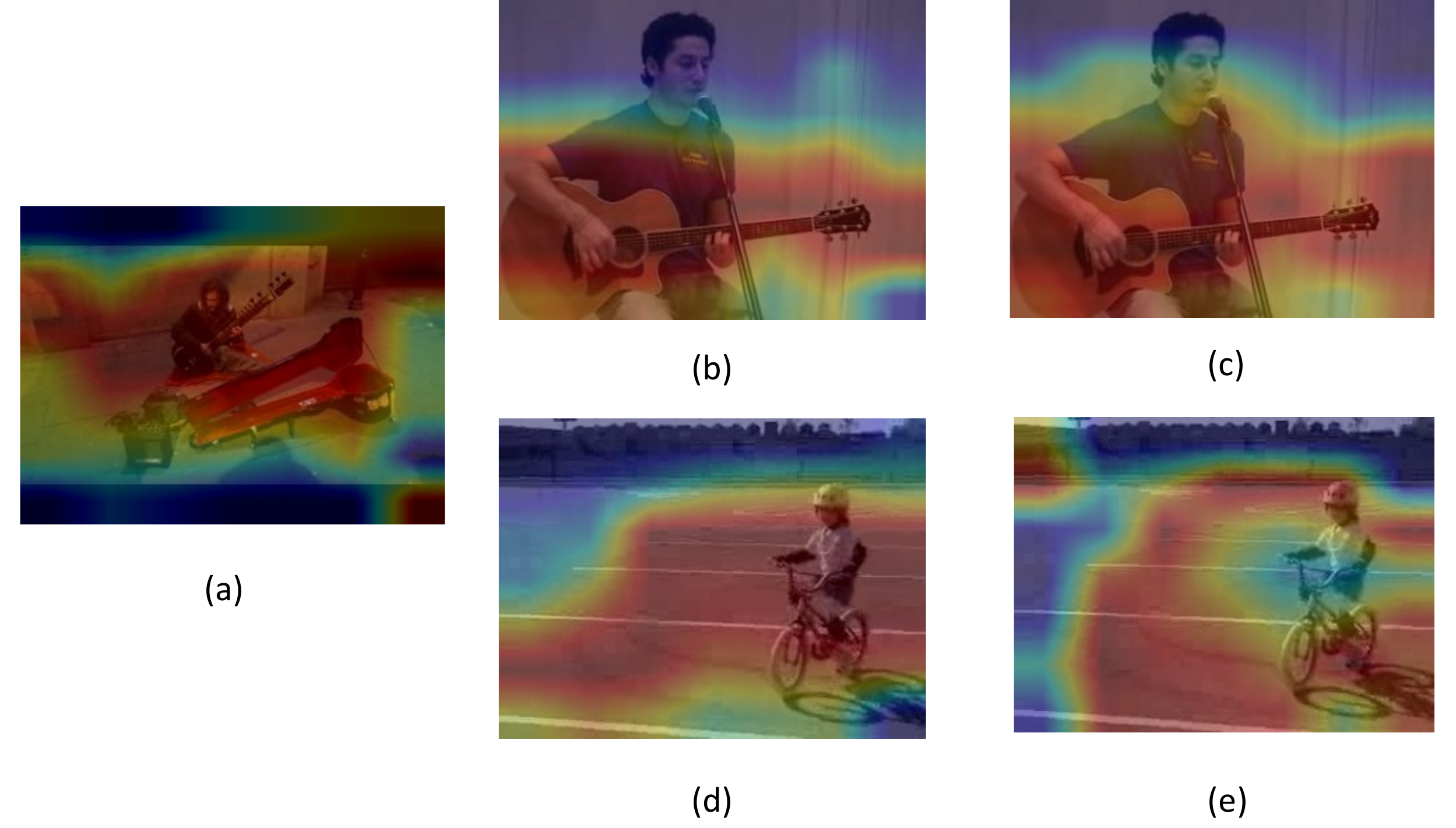}
\end{center}
\caption{Heatmaps showing activations of various classes' classifier layers on various class videos. (a) is the display of the activation from the ``playing sitar" class on a``playing sitar" video, (b) is the display of the activation from the ``playing guitar" class on a``playing guitar" video, (c) is the display of the activation from the ``playing sitar" class on a``playing guitar" video, (d) is the display of the activation from the ``biking" class on a``biking" video and (e) is the display of the activation from the ``playing sitar" class on a``biking" video. These heatmaps show that test class ``playing sitar" is correctly learning from training class ``playing guitar" instead of training class ``biking"}
\label{fig:heatmaps}
\end{figure}

To understand the impact of using \texttt{KG1}, \texttt{KG2} and \texttt{KG3} for learning each test class, we plot the class-wise accuracy for UCF101 and HMDB51 in Fig.~\ref{fig:bar_plot}. Each color of the bar represents a different KG: blue is for word based \texttt{KG1}, orange is for visual feature based \texttt{KG3} and grey is the combination of \texttt{KG1}, \texttt{KG2} and \texttt{KG3}.

As observed in Fig.~\ref{fig:bar_plot}, for few classes such as ``billiards'', ``talk'', ``playing tabla'', \texttt{KG1} performs the best. These classes innately have many neighbors in the word embeddings space, which help in learning them from given training classes. Few other classes, such as ``front crawl swimming'', ``pommel horse gymnastics'', ``chew food'' and ``pour liquid'' perform well with just \texttt{KG1} as well, since we add the extra word ``swimming'', ``gymnastics'', ``food'' and ``liquid'' respectively, to enforce good neighbors in language domain. Intuitively, \texttt{KG3} does well for ``uneven bars'', ``fall floor'', ``smile'' and ``shoot gun'', since these have distinct visual features. The combination KG works well for ``still rings'', ``parallel bars", ``jumping jack", ``playing dhol'', ``climb stairs'', ``talk" and ``wave''.







\paragraph{\bf Ablation for Network Architecture:}

We experiment with different number of layers of the GCN (2,4 6, 8 and 10) to explore influence of GCN depth on performance for both UCF101 and HMDB51. The increase in the number of layers of the GCN increases smoothing and decrease in number of layers causes less information propagation. We found that 6 layers gives us the best performance. 

\paragraph{Usefulness of GCN vs a linear combination of training class weights:}
To show the performance improvement due to GCN compared to just linear combinations, we perform an ablation study.  For each test class, we find the top 4 neighbors in the training set. Then using the adjacency edge connection weights, the classifier layer weight for the test class is a weighted average of the classifier layer weights for its neighbors. The performance is in Table~\ref{table:linear_combo}.

\paragraph{Use encoder decoder before GCN:}
We run another set of experiments where a 2 layered encoder decoder network is added before GCN, for improving encoding of sentence embedding features. The results do not show any promise as seen in Table~\ref{table:enco_deco}.

\paragraph{\bf Random test train splits:}
Some of the experiments are done on a random sub-sample of the test-set classes. For UCF101, we choose 10 out of 23 classes 5 times; so that for each random sample of 10 test classes, the rest of the 91 classes forms the training set. The mean accuracy score is calculated after each run and the result of all 5 runs are averaged to get the final mean accuracy score. The results for each of these splits is in Table~\ref{table:results1}.

\paragraph{\bf Learning classifier for unknown classes from related
classes in Knowledge Graph:}
The heatmaps in Figure~\ref{fig:heatmaps} depicts the test nodes learning from the interconnections to the train nodes in the KG. They are based on CAM~\cite{zhou2016learning}.
Considering the test class ``playing sitar” in UCF101, one of the top 5 nearest train classes in UCF101 is “playing guitar” and one of the random classes that have no relation is “biking”. Now among the five sub-figures in Figure~\ref{fig:heatmaps},
(a) is the display of the activation from the ``playing sitar” class on a ``playing sitar” video, (b) is the display of the activation from the ``playing guitar” class on a ``playing guitar” video, (c) is the display of the activation from the ``playing sitar” class on a“playing guitar” video, (d) is the display of the activation from the ``biking” class on a ``biking” video and (e) is the display of the activation from the ``playing sitar” class on a ``biking” video. What we show here is that
``playing sitar” classifier is similar to the ``playing guitar” classifier and hence the
heat maps from both are similar. This is not the case between ``playing sitar”
and ``biking”.


\section{Conclusion}

In this work we investigate different combinations of knowledge graphs (KG) for actions that give better performance for zero and few shot action recognition. We show significant improvement on zero shot learning by using a network that models a sequence of words instead of traditional single word based models. Moreover, extending KG using other action classes leads to better results. We observe that combining word based knowledge graphs with visual knowledge graphs help in few shot learning. Also combining verbs and noun based KG, improves both zero and few shot learning. 
Work on dynamically learning the graph weights can be explored in the future.

\begin{acknowledgements}
This work was supported by the Air Force, via Small Business Technology Transfer (STTR) Phase I (FA8650-19-P-6014) and Phase II (FA864920C0010), and Defense Advanced Research Projects Agency (DARPA) via ARO contract number W911NF2020009.


\end{acknowledgements}

%
%

\bibliographystyle{apalike}
\bibliography{biblio}   



\end{document}


\sloppy
\title{All About Knowledge Graphs for Actions - Supplementary
}


\author{Pallabi Ghosh$^1$        \and
        Nirat Saini$^1$\and
        Larry S. Davis$^1$ \and
        Abhinav Shrivastava$^1$
}
\institute{$^1$Department of Computer Science, University of Maryland, College Park, MD, USA\\
\email{\{pallabig, nirat, lsd, abhinav\}@cs.umd.edu}}

\authorrunning{Ghosh et al.} 


\date{Received: date / Accepted: date}

\maketitle

\section{Details about Datasets}

\begin{table}
\caption{Sample noun and verb detected from action class names for UCF101, HMDB51 and Kinetics}
\label{table:nounverb}
\begin{center}
\begin{tabular}{@{}lll@{}}
\hline
\multicolumn{3}{c}{UCF101} \\ 
Action & Noun & Verb \\ 
\hline
playing sitar & sitar & playing \\ 
playing tabla & tabla & playing \\ 
basketball dunk & basketball & dunking \\ 
\hline
\multicolumn{3}{c}{HMDB51} \\
Action & Noun & Verb \\ 
\hline
brushing hair & hair & brushing \\
clapping & applause & clapping \\
pullup & pullup & doing \\
\hline
\multicolumn{3}{c}{Kinetics} \\
Action & Noun & Verb \\ 
\hline
applying cream & cream & applying \\
archery & archery & doing \\
arm wrestling & arm & wrestling \\
\hline
\end{tabular}
\end{center}
\end{table}

\begin{table}
\caption{Some samples of classes in UCF101 and HMDB51 that are common with Kinetics and hence cannot be in the test sample for zero-shot learning using models pre-trained on Kinetics}
\label{table:similar_classes}
\centering
\resizebox{\linewidth}{!}{
\begin{tabular}{@{}llll@{}}
\hline
\multicolumn{4}{c}{\bf{UCF101}} \\ 
\bf{UCF101} & \bf{Kinetics} & \bf{UCF101} & \bf{Kinetics} \\ \hline
archery & archery & boxing & punching \\
& & punching bag & bag\\ 
\hline
baby & crawling & boxing & punching \\ 
crawling &  baby &  speed bag &  bag \\ 
\hline 
band & marching & breast stroke & swimming \\ 
marching & &  & breast stroke\\
\hline
baseball & catching or & brushing & brushing \\ 
pitch & throwing baseball &  teeth & teeth \\ 
\hline
basketball & dribbling & clean and & clean and \\ 
& basketball &  jerk &  jerk \\ 
\hline
basketball& dunking & cliff& diving \\ 
 dunk & basketball &  diving  & cliff  \\ 
\hline
bench & bench & cricket & playing \\ 
press &  pressing &  bowling & cricket \\ 
\hline
biking & biking  & cricket & playing \\ 
& through snow &  shot &  cricket \\ 
\hline
haircut & getting a & cutting & cutting  \\ 
 &  haircut & in kitchen & pineapple  \\ 
 & & & cutting  \\ 
 & & &  watermelon \\ 
\hline
blowing & blowing & diving & scuba diving \\ 
candles & out candles & & springboard \\ 
& & & diving \\ 
\hline
bowling & bowling & drumming & playing \\ 
& & &  drums \\ 
\hline
\hline
\\
\multicolumn{4}{c}{\bf{HMDB51}}\\
\bf{HMDB51} & \bf{Kinetics} & \bf{HMDB51} & \bf{Kinetics} \\
\hline
brush hair & brushing hair & cartwheel & cartwheeling \\
\hline
eat & eating burger & clap & clapping\\
& eating cake & & \\
\hline
Climb & climbing ladder & golf & golf driving\\
& climbing tree & & golf chipping \\
\hline
dribble & dribbling basketball & drink & drinking \\ 
\hline
dive & springboard diving & hit & hitting \\
& scuba diving & & baseball \\
\hline
\end{tabular}}
\end{table}

We give some sample noun and verb classes that we use to construct KG2 in Table~\ref{table:nounverb}. We don't show samples for Charades because the dataset provides nouns and verbs.
We also give some sample classes in UCF101 and HMDB51 that are in common with Kinetics and had to be removed from test set in Table~\ref{table:similar_classes}. Note that the names always do not exactly match, but they are either the same class or they contain similar videos. 
Finally the test classes in each of the 3 datasets, UCF101~\cite{soomro2012ucf101}, HMDB51~\cite{kuehne2013hmdb51} and Charades~\cite{charades2016} that are not in common with Kinetics~\cite{kay2017kinetics} are in Table~\ref{table:test_classes_ucf_hmdb_kinetics}.

\begin{table*}
\caption{Test classes from UCF101, HMDB51 and Charades that are not in common with Kinetics Dataset}
\label{table:test_classes_ucf_hmdb_kinetics}
\begin{subtable}
\begin{center}
\begin{tabular}{lllll}
\hline
\multicolumn{5}{c}{UCF101 test classes}\\
\hline
apply eye makeup & apply lipstick & balance beam & billiards & field hockey penalty \\
front crawl & hammering & handstand walking & jumping jack & mixing batter \\
nunchucks & parallel bars & playing daf & playing dhol & playing sitar \\
playing tabla & pommel horse & rafting & still rings & table tennis shot \\
typing & uneven bars & yo yo  &  &\\
\hline
\end{tabular}
\end{center}
\end{subtable}

\begin{subtable}
\vspace{-0.2cm}
\begin{center}
\begin{tabular}{llllll}
\hline
\multicolumn{6}{c}{HMDB51 test classes} \\
\hline
chew & climb stairs & fall floor & handstand & pour & shoot gun\\
sit & smile & stand & talk & turn & wave \\
\hline
\end{tabular}
\end{center}
\end{subtable}

\begin{subtable}
\vspace{-0.2cm}
\begin{center}
\begin{tabular}{lll}
\hline
\multicolumn{3}{c}{Charades test classes} \\
\hline
closing a box & closing a closet/cabinet & closing a laptop \\

closing a refrigerator & closing a window & fixing a door \\
 
fixing a doorknob & fixing a light & holding a blanket \\

holding a box & holding a broom & holding a cup of something \\

holding a dish & holding a laptop & holding a phone \\

holding a shoe/shoes & holding a towel/s & holding a vacuum\\

holding some medicine & lying on a bed & opening a closet \\

playing with a phone & putting a bag somewhere & putting a blanket somewhere \\

putting a dish/es somewhere & putting a laptop somewhere & putting a phone somewhere \\

putting a picture somewhere & putting a towel/s somewhere & putting groceries somewhere \\

putting something on a shelf & reaching for and grabbing a picture & sitting at a table \\

sitting in a bed & sitting on a table & sitting on sofa \\

smiling in a mirror & snuggling with a pillow & someone is awakening somewhere \\

someone is dressing & someone is holding a paper & someone is smiling \\

someone is standing up from somewhere & someone is undressing & taking a bag from somewhere \\

taking a blanket from somewhere & taking a box from somewhere & taking a broom from somewhere \\

taking a cup from somewhere & taking a dish/es from somewhere & taking a laptop from somewhere \\

taking a phone from somewhere & taking a picture of something & taking a towel/s from somewhere \\

taking a vacuum from somewhere & taking paper from somewhere & taking something from a box \\

taking some medicine & throwing a blanket somewhere & throwing a broom somewhere \\

throwing food somewhere & tidying a shelf or something on a shelf & tidying up a closet \\

tidying up a towel/s & tidying up with a broom & turning off a light \\

turning on a light & walking through a doorway & wash a dish \\

washing a mirror & washing a window & watching a laptop or something on a laptop \\

watching something in a mirror & watching television & watching at a picture \\

watching outside of a window & working on paper & working on a laptop \\
\hline
\end{tabular}
\end{center}
\end{subtable}
\end{table*}

\bibliographystyle{apalike}
\bibliography{biblio}   
